\documentclass[letterpaper, 10pt, conference]{ieeeconf}      % Use this line for a4 paper

\IEEEoverridecommandlockouts                              % This command is only needed if 
\makeatletter
\let\NAT@parse\undefined
\makeatother

\usepackage{amsmath} % assumes amsmath package installed
\usepackage[american]{babel}
\usepackage{graphicx}
\usepackage{xcolor}
\usepackage{easy-todo}
\usepackage[numbers]{natbib}
\usepackage{siunitx}
\usepackage{hyperref}

\usepackage{pgfplots}

\DeclareMathOperator*{\argmax}{arg\!\max}

\title{\LARGE \bf
Deep Detection of People and their Mobility Aids for a Hospital Robot
}

\author{Andres Vasquez \hspace{0.5cm} Marina Kollmitz  \hspace{0.5cm} Andreas Eitel \hspace{0.5cm} Wolfram Burgard% <-this % stops a space
\thanks{*This work has been partially supported by the German Federal Ministry of  Education and Research (BMBF), contract number 01IS15044B-NaRKo and by the DFG
under grant number EXC 1086.}% <-this % stops a space
\thanks{The authors are with the Faculty of Computer Science,
        University of Freiburg, Freiburg, Germany. Corresponding author: 
        {\tt\small kollmitz@informatik.uni-freiburg.de}}%
}

\begin{document}

\maketitle
\thispagestyle{empty}
\pagestyle{empty}

%%%%%%%%%%%%%%%%%%%%%%%%%%%%%%%%%%%%%%%%%%%%%%%%%%%%%%%%%%%%%%%%%%%%%%%%%%%%%%%%
\begin{abstract}

  Robots operating in populated environments encounter many different types of people, some of whom might have an advanced need for cautious interaction, because of physical impairments or their advanced age. Robots therefore need to recognize such advanced demands to provide appropriate assistance, guidance or other forms of support.  In this paper, we propose a depth-based perception pipeline that estimates the position and velocity of people in the environment and categorizes them according to the mobility aids they use: pedestrian, person in wheelchair, person in a wheelchair with a person pushing them, person with crutches and person using a walker.  We present a fast region proposal method that feeds a Region-based Convolutional Network (Fast R-CNN~\cite{girshick2015fast}). With this, we speed up the object detection process by a factor of seven compared to a dense sliding window approach. We furthermore propose a probabilistic position, velocity and class estimator to smooth the CNN's detections and account for occlusions and misclassifications. In addition, we introduce a new hospital dataset with over 17,000 annotated RGB-D images. Extensive experiments confirm that our pipeline successfully keeps track of people and their mobility aids, even in challenging situations with multiple people from different categories and frequent occlusions. Videos of our experiments and the dataset are available at \url{http://www2.informatik.uni-freiburg.de/~kollmitz/MobilityAids}.

\end{abstract}

%\newcommand{\etal}{\textit{et al.}} %added by andres --> not necessary, just use \citet{...} command

%%%%%%%%%%%%%%%%%%%%%%%%%%%%%%%%%%%%%%%%%%%%%%%%%%%%%%%%%%%%%%%%%%%%%%%%%%%%%%%%
\section{INTRODUCTION}

Mobile robots operating in populated environments need to perceive and react to the people they encounter. Our research is part of the project NaRKo, which aims at employing autonomous robots for delivery tasks as well as for guiding people to treatment rooms in hospitals. Hospital environments pose special challenges to autonomous robot operation, because the people interacting with the robot might have very different needs and capabilities. It is therefore desirable for the robot to adapt its behavior accordingly, e.g. by adjusting its velocity and path when guiding a person with a walking frame compared to a healthy person without motion impairments. 

\begin{figure}[t]
\centering
\includegraphics[width=\columnwidth]{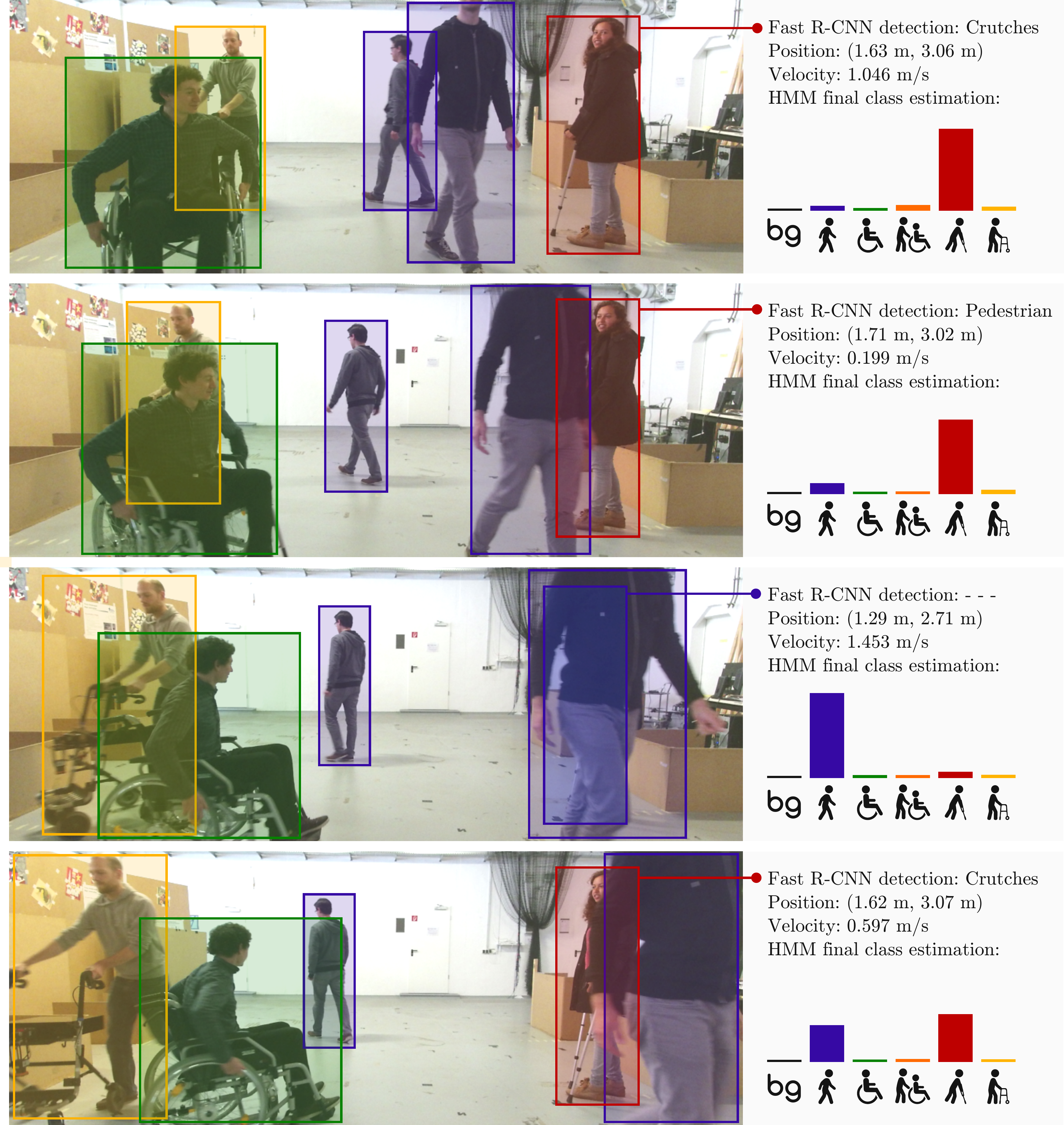}\\
% \caption{Our perception pipeline employs the Fast R-CNN detection method to estimate the position, velocity and category of people in populated environments. We exploit the ability of the Kalman filter to do predictions to estimate the position and velocity in absence of observations. The category of a person is estimated in a probabilistic fashion by a hidden Markov model.}
% \caption{We present a perception pipeline, that estimates the position, velocity and the class of people, recorded in different populated environments including a hospital. We employ a region-based convolutional network for object detection and combine it with a tracking module for estimation of position and velocity. Further, we filter the class of a person over time using a probabilistic model.}
\caption{We present a depth-based perception pipeline that estimates the position, velocity and the class of people, recorded in different populated environments, including a hospital. Top: a person with crutches is detected at the right corner of the image and the class estimation predicts high probability for the respective class. Second and third: the person is occluded by another person and the estimated class switches gradually to pedestrian, while the tracker keeps the track alive. Bottom: the class estimator reflects the ambiguity of the belief about the class of the person. 
We show RGB images for visualization purposes only.
}
\label{fig:cover1}
\end{figure}

\begin{figure*}[t]
\centering
\includegraphics[width=13cm]{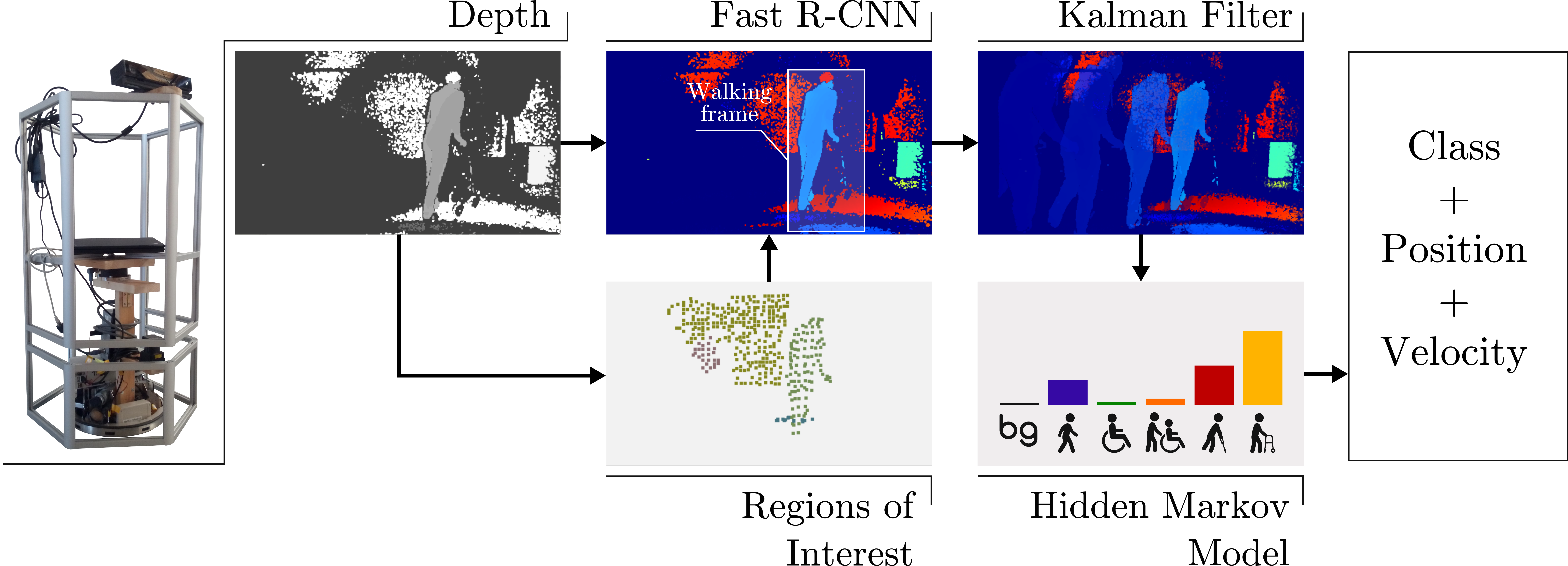}\\
%\caption{ Pipeline proposed to estimate class, position and velocity of people in hospital environments.}
\caption{Our pipeline operates on depth images collected from a Kinect v2 sensor. To achieve a fast runtime we present a depth-based region proposal method that generates regions of interest based on Euclidean clustering, which we feed into a Fast R-CNN detection network. We filter the resulting bounding box detections using a Kalman filter for position and velocity estimation of people. Further, we employ a hidden Markov model for filtering the class predictions over time.}
\label{fig:pipeline1}
\end{figure*}

Our work targets the detection and categorization of people according to the mobility aids they use. Privacy concerns play an important role in the hospital, which is why our approach is based only on depth data. However, depth data conveys a lot less information than RGB images, which makes the problem more challenging.
We propose a perception pipeline which uses depth images from a Kinect v2 camera at 15 frames per second and outputs the perceived class, position, velocity and the tracked motion path of people. Our object detection pipeline uses the Fast R-CNN method proposed by \citet{girshick2015fast},
which takes an image together with a set of regions of interest (ROIs) as its input and outputs classification scores for each ROI. We propose a fast depth-based ROI proposal method that uses ground plane removal and clustering to generate a set of regions and applies a set of local sliding templates over each region. We compare our method against a dense sliding window baseline and show that our approach is significantly faster and yields improved performance. 
The perceived class of each ROI as well as its position in the world frame are further processed by our probabilistic position, velocity and class estimator. In addition to tracking the position and velocity of each person in the environment with a Kalman filter, we use a hidden Markov model (HMM) to estimate the class of each track.
As depicted in Fig.~\ref{fig:cover1}, the probabilistic position, velocity and class estimator resolves occlusions and outputs a probability distribution over the five classes, taking the previous observations into account.  
%Neural network outputs are often interpreted as probabilities over the output classes, but they are mostly trained to produce peaked output distributions, resulting in over-confident estimations. Instead, our class estimator uses observation and transition probabilities obtained from data and takes previous observations into account, which results in smoothed output distributions. 
This paper further presents our hospital dataset that contains over 17,000 annotated RGB-D images with 960x540 pixel resolution. 
%It is publicly available at \url{http://www2.informatik.uni-freiburg.de/~kollmitz/MobilityAids}. 
We collected the dataset in the facilities of the Faculty of Engineering of the University of Freiburg and in a hospital in Frankfurt. The webpage also shows a video of the final results of our approach.

\section{RELATED WORK}
%% People detection and tracking from mobile platforms %%
People detection and tracking is a widely studied field in both computer vision and robotics. Given the large amount of previous work in this area, we will focus only on approaches that integrate both people detection and tracking in a combined system.
Further special emphasis is laid on approaches for mobile platforms that are equipped with vision-based sensors such as cameras or RGB-D sensors.

\citet{ess2009} address the problem of multi person tracking and detection using a stereo vision system mounted on a mobile platform, integrating visual odometry, depth estimation and pedestrian detection for improved perception.~\citet{choi2013} propose a method to track multiple people from a moving platform based on a particle filter approach. Several different detectors are used such as upper body, face, depth-based shape, skin color and a motion detector. 
%A different tracking ~\citet{mitzel2012} propose a tracking-before-detection strategy , using an ICP-based tracking scheme to create a 3D model which is fed into a classifier. As a side product of the model learning, the method can detect items that people are carrying which appear as outliers in the model. 
%Similar to ours, several framerworks exist, that take as input RGB-D images and output tracks of people in real-time~\cite{dondrup2015,jafari2014real, munaro2014}. 
Recently, extensive frameworks that include several people detection and tracking methods for mobile robots operating in indoor environments have been presented~\cite{linder2016,dondrup2015}. In comparison to the mentioned frameworks, we focus on a multi-class detection problem and do not only track position and velocity but also the class throughout time. Further, previous approaches rely on manually designed detectors for different body parts while we use a single neural network detector that learns those body features automatically. 

% Multi-Class object detection
Our work is further related to the research area of object detection, which recently is dominated by deep neural network approaches, most prominently by region-based convolutional neural networks~\cite{girshick2015fast,renNIPS15fasterrcnn}, which achieve very good results but are not applicable for real-time yet. An interesting extension to the region-based CNN detection approaches is the very recently introduced region-based fully convolutional neural network presented by~\citet{li2016r}, which increases the test-time speed.
Recently~\citet{redmon2016yolo9000} proposed an approach that formulates object detection as a regression problem. It can operate in real-time and achieves very impressive performance on several object detection benchmarks.  
We also employ a region-based convolutional neural network classifier and, to achieve a fast runtime, we combine it with our depth-based region proposal method. 
Recent work on multi-class object recognition and detection applied to mobile robot scenarios include a Lidar-based wheelchair/walker detector~\cite{beyer2017drow} and a human gender recognition approach~\cite{linder2015real}. To the best of our knowledge there exists no prior work that presents multi-class people detection applied to service robot scenarios.

%% region proposals
Region of interest (ROI) extraction is often used to speed up the detection process and to reduce the number of false positives. Often employed is the assumption that most objects occur on a dominant ground surface and several methods exist that generate ROIs based on this ground plane assumption~\cite{munaro2014,spinello2008multimodal,sudowe2011efficient}.  \citet{munaro2014} detect and remove the ground plane, then they apply Euclidean clustering. To overcome the problem of having two or more people in the same cluster they run an additional head detector. Our method is similar, although we use a local sliding window approach with templates that are applied to each cluster instead of a head detector.  The people detector of \citet{jafari2014real} extracts ROIs by fusing point clouds over a time window to compute a 2D occupancy map. \citet{spinello2011people} also exploit depth information to reduce the number of candidates in a sliding window approach for people detection in RGB-D data. However, their approach is sensitive to false depth readings that can result in very large proposal windows. \citet{chen20153d} present a more evolved method to generate ROIs formulating an energy function that encodes informative features such as object size priors, ground plane and depth information. Despite good performance, the algorithm has a computation time of approximately one second. Our ROI extraction approach instead is fast and simple.
%% other datasets 

Another contribution of this paper is a novel, annotated large-scale dataset for multi-class people detection.
In the literature there are several other datasets that include multi-class labels for people, mostly from the area of human attribute recognition~\cite{sharma2011learning,bourdev2011describing} or more specifically gender recognition~\cite{linder2015real}. Our dataset can be valuable for the robotics community, because on one hand it provides a large number of labeled images and on the other hand it is recorded from a mobile platform. Very recently and most similar to our dataset \citet{sudowe2015person} recorded video sequences of people from a moving camera for the task of human attribute recognition. 

\section{PEOPLE PERCEPTION FRAMEWORK}
Fig.~\ref{fig:pipeline1} gives an overview of our overall system, which takes as input a stream of depth images, computes a set of ROIs, classifies those proposals and filters the network output over time. Specifically we filter the position and velocity of a person using a Kalman filter and their category using a hidden Markov model.
The resulting output of our framework, which contains the class, position and velocity of people, is visualized in Fig.~\ref{fig:cover1}.

\subsection{Fast Depth-Based Region Proposals Generation}

To obtain the regions of interests we follow a sequence of steps. After converting the depth image into a point cloud representation
we remove all points belonging to the ground plane, apply Euclidean clustering and slide a set of local sliding templates over the obtained segments, see Fig.~\ref{fig:ROI}.

\subsubsection{Ground plane removal}
For a fast ground plane estimation we apply Random Sample Consensus to estimate the parameters of the ground plane.
After the removal of the ground plane we segment the point cloud by means of the Euclidean clustering method. %Finally we project bounding boxes around the centroid of each of these segments. 
\subsubsection{Local sliding templates}
\begin{figure}[t]
\centering
\includegraphics[width=6.5cm]{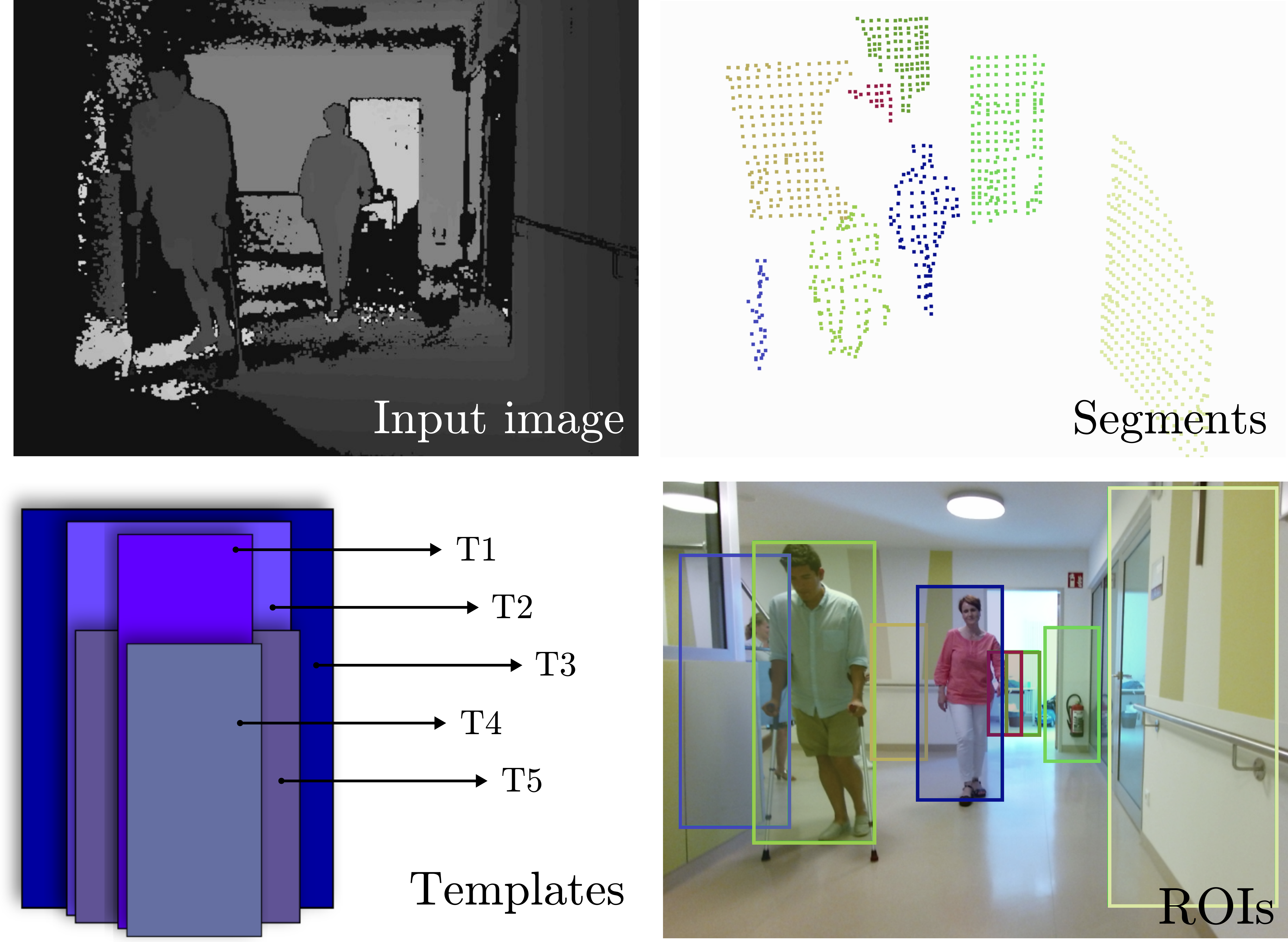}\\
\caption{We obtain regions of interest (ROIs) from depth images, based on point cloud segmentation and the use of templates. For demonstration purposes, we only show the template $T1$ applied to each segment in the lower right image.}
\label{fig:ROI}
\end{figure}

People of the five classes have different outlines in the images, because of the different typical shapes. A pedestrian, for example, takes up a rather narrow and tall part while a person in a wheelchair will take up a wider and lower part. To account for the different contours we use five different local sliding templates as shown in Fig.~\ref{fig:ROI}. All five templates are projected around the centroid of each computed point cloud segment to compute the ROIs.
%For each segment, we need to compute ROIs of different sizes and aspect ratios, given the variety of object classes we want to detect. In order to cover this variation we use five different local sliding templates, see Fig.\ref{fig:ROI}. 
The first bounding box considers the size of an average pedestrian with height $h_p=1.75m$ and width $w_p=0.4m$, this rectangle is our template box $T_1$ for detecting pedestrians and the front views of the other categories except people using wheelchairs.  
We stretch $T_1$ horizontally with a factor of $1/3$ to each side to obtain the template box $T_2$ for side views of people using crutches or walking frames. In the same way, $T_1$ is stretched again with a factor of $2/3$ to obtain $T_3$ for the side view cases of people pushing people in wheelchairs. Finally, we obtain the template boxes $T_4$ and $T_5$ for people using wheelchairs by reducing the height of $T_1$ and $T_2$ by a factor of $1/4$.

Note that a candidate in the point cloud might contain only a part of a human body and since we do not have this prior information, we take into account that the centroid of a segment is not always the center of a body. Therefore, we horizontally slide these five proposals to $l$ different positions around the center of the candidate, using a stride of $n_s$ pixels. Accordingly, we end up with $5l$ proposals for every segment that will be fed into the Fast R-CNN detector.  
\subsection{Detection using Fast R-CNN}
%The classification task is solved by means of the Fast Region Based Convolutional Neural Network (Fast R-CNN) introduced by %\citet{girshick2015fast}. 
The input of the Fast R-CNN network is a color-encoded depth image together with a set of ROIs and its output are both softmax scores for each ROI and bounding box coordinates from a regression layer.
We use the deep network architecture proposed by~\citet{mees2015choosing}, which contains 21 convolutional and six max pooling layers (GoogLeNet-xxs).

\subsubsection{Training}
Fast R-CNN jointly optimizes a multi-task loss that contains a softmax classifier and a bounding box regressor. 
%For training we use stochastic gradient descent (SGD) with momentum and the hierarchical data sampling procedure provided by the authors. More specifically for each minibatch we sample $N$ images and then $R/N$ ROIs from each image, using $N=2$ and $R=128$.      
We build the set of ROIs for the training stage by applying a dense multi-scale sliding window approach using our five template bounding boxes previously described ($T1,\dots,T5$). We slide them at different scales all over the image, which produces a set of dense ROIs containing more than 29'000 bounding boxes for a single image, from which we sample during training. 
For every sampled ROI, we compute the overlap with a ground-truth bounding box. This overlap we measure in terms of Intersection over Union (IoU). A sample with IoU greater than 0.6 is considered a positive training example of the class contained in the ground-truth bounding box. Samples with IoU below 0.4 are instances of the background class. We run stochastic gradient descent SGD for 140,000 iterations, using a learning rate of 0.001 and momentum of 0.9.

\subsubsection{Test-time detection}

Fast R-CNN outputs scores $s^r_m$ for each class $m \in \{1,\ldots,M\}$ and each proposal $r$. The class $c^r$ corresponding to each proposal is the one with the highest score:
\begin{equation}
c^r=\underset{m=1}{\overset{M}{\argmax}}~s^r_m
\end{equation}
Since we manage $5l$ different proposals (local sliding templates) for the same candidate segment, we will have several proposal classes $c^r$ for one segment. We assign the final class of the segment as the class with the highest number of appearances for all associated proposals.
Note that this procedure assumes that each segment contains not more than one person, and in practice this assumption works reasonably well. 
%A segment will be assigned the background class only if all templates where detected as background. 
%The bounding box containing the classification is selected as the one with the highest score from the boxes of the class %assigned to the segment.   
Two segments corresponding to the same person or very close segments in the point cloud might result in overlapping detections. We overcome this problem by applying non-maximum suppression (NMS) as a final step. NMS chooses a subset of the remaining $c^r$, which is the final output $c$ of the Fast R-CNN detector. The corresponding position $(x_c, y_c, z_c)$ of the person in camera coordinates is obtained from the bounding box coordinates, which are also provided by the Fast R-CNN network, and the distance of the segment from the camera.

\subsection{Probabilistic Position, Velocity and Class Estimation}

The detection stage provides us with a set of coordinates for each bounding box containing a person of the form $(x,y,z,c)$, where $(x,y,z)$ is the detected pose of the person transformed from camera coordinates into the world frame and $c$ is the perceived class. The world frame transformation requires knowledge of the robot's position in the environment, which we obtain from its odometry. Our position, velocity and class estimator computes the belief $\text{Bel}(\mathbf{x})$ of the person state $\mathbf{x}=(x_x, x_y, x_{\dot{x}}, x_{\dot{y}}, x_c)$, where $(x_x,x_y)$ represent the person's true position and $(x_{\dot{x}},x_{\dot{y}})$ the true velocity on the ground plane and $x_c$ represents the true class. %To this end, the estimator combines two modules: A Kalman filter-based multi-tracking module estimating the position and velocity for all observed people and a hidden Markov model (HMM) estimating the class of each person. 

Each person has one Kalman filter and one HMM associated to it. The Kalman filter uses a constant velocity motion model, where the motion and the observation noise are obtained experimentally by analyzing the labeled ground truth trajectories and the corresponding Fast R-CNN detections of people in the training data set. We solve the data association problem between frames using the Mahalanobis distance 
\begin{eqnarray}
d_{ij}^2 &=& v_{ij}^T(t) \hat{S}^{-1}(t) v_{ij}(t) \\
\textnormal{with} \hspace{0.7cm} v_{ij} &=& z_i(t) - H(t)\hat{x}_j(t) \\ 
\textnormal{and} \hspace{0.5cm} \hat{S}(t) &=& H(t)\hat{P}(t)H^T(t) + R(t),
\end{eqnarray}
where $\hat{x}(t)$ and $\hat{P}(t)$ are the predicted state mean and covariance at time $t$, $z(t)$ is the observation, $H(t)$ the observation model and $R(t)$ the observation noise of the filter. The observation and filter indices at time $t$ are $i$ and $j$. We use the Hungarian algorithm to assign tracks to observations, according to the pairwise Mahalanobis distances. If the distance is larger than a threshold, the observation is not paired to a track. Instead, a new Kalman filter is initialized. If there is no observation for a track, we perform a prediction without observation update. 

Each Kalman filter has one HMM associated to it for estimating the class $x_c$ of the tracked person, according to
%The class $x_c$ of a person is estimated by a hidden Markov model according to
\begin{equation}
\begin{aligned}
p(x_{c,t}&\mid c_{1:t}) = \eta p(c_{t} \mid x_{c,t}) \\ 
& \sum_{x_{c,{t-1}}} p(x_{c,t} \mid x_{c,t-1}) p(x_{c,t-1} \mid c_{1:t-1}).
\end{aligned}
\label{equ:hmm1}
\end{equation}
Eq.~\ref{equ:hmm1} models the probability of the tracked person to belong to a given class $x_{c,t}$, given the past observations $c_{1:t}$. Here, $x_{c,t}$ is hidden, since we only get measurements $c_t$ for it. The measurement model $p(c_{t} \mid x_{c,t})$ connects the hidden with the observed variable for time step $t$. The HMM further assumes that the class $x_{c,t}$ can randomly change from one time step to the next, represented by the transition model $p(x_{c,t} \mid x_{c,t-1})$. In a hospital, a possible transition could be person with crutches $\rightarrow$ pedestrian $\rightarrow$ person in wheelchair for a patient who has just finished physiotherapy and hands over his crutches to return to his wheelchair. We need to further specify the class prior $p(x_{c,1})$ for the initialization of the HMM. 

The output of deep neural network classifiers like Fast R-CNN can be interpreted as $p(x_{c,t}\mid c_t)$, since it represents a probability distribution.
%like summing up to one. 
However, training with one-hot encoded labels results in very peaky distributions and over-confident estimates. Therefore, we will not employ the network scores $s^r_m$ directly for our HMM. Instead, we analyze our training data to determine the underlying probability distributions statistically. To this end, we first generate and label all proposals for each frame in the validation set of our data, given by our ROI generator and obtain a classification output for each ROI. The percentage of labels for each class determines the class prior $p(x_{c,1})$. The measurement model $p(c_t\mid x_{c,t})$ is determined by the amount of detections for each class, given a certain label. As a side node, the measurement model corresponds to the classifier's confusion matrix. The transition model $p(x_{c,t} \mid x_{c,t-1})$ is given by the amount of transitions from one class to the other with respect to the total number of transitions.

Due to the limited amount of examples in the validation set, we might not observe all class transitions, even if they are possible. 
Therefore, we assign small probabilities to all unobserved but possible transitions and misdetections, using a Dirichlet prior. The data association for the HMM is given by the Kalman filter. If a tracked person is in the field of view of the camera and there is no observation $c_t$ for time step $t$, we treat it as a background detection in the HMM. If the track is outside the field of view, we apply the transition model to the previous state estimate without considering an observation.
%We maintain one combined Kalman Filter and HMM for each observed person to recursively estimate their position, velocity and class. 
The position, velocity and class estimator removes tracks with a standard deviation in position above a threshold. Furthermore, tracks which are estimated as background with a probability above a threshold are deleted.

\section{EXPERIMENTS}
In this section we evaluate the performance of the two submodules Fast R-CNN object detection and Kalman filter tracking. Additionally, we present quantitative results for the combined perception pipeline. For these experiments we used our hospital dataset. Finally, we present a real-robot scenario where the robot uses our perception pipeline in order to give special assistance in a person guidance task.
%Due to privacy issues in some scenarios we cannot use RGB information and only depth information is available. This realistic constraint especially in a hospital environment motivated us to present all quantitative results using only depth data.

\subsection{Dataset}

In order to train and evaluate our pipeline we annotated our hospital dataset. Images were collected in the facilities of the Faculty of Engineering of the University of Freiburg and in a hospital in Frankfurt using a mobile Festo Robotino robot, equipped with a Kinect v2 camera mounted \SI{1}{\meter} above the ground and capturing images at 15 frames per second. The robot was controlled by a notebook computer running ROS (Robot Operating System). 

\begin{figure}
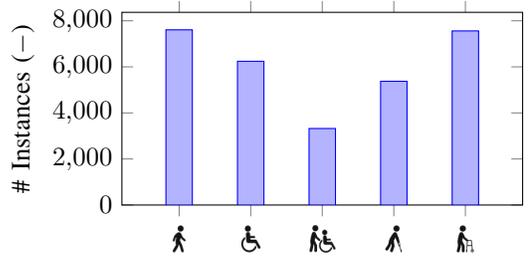

\centering
\begin{tikzpicture}
\begin{axis}[
legend columns = -1,
legend style={
at={(0.5,1.03)},
anchor=south},
ybar, 
ymin=0,
width = 0.8\columnwidth,
height = 0.48\columnwidth,
bar width = 10pt,
ylabel = {\# Instances ($-$)},
ylabel near ticks,
%nodes near coords,
symbolic x coords={Pedestrian, Wheelchair, Push_wheelchair, Crutches, Walker}, 
xticklabels={\includegraphics[width=0.35cm]{icon_pedestrian}, \includegraphics[width=0.35cm]{icon_wheelchair}, \includegraphics[width=0.35cm]{icon_push_wheelchair}, \includegraphics[width=0.35cm]{icon_crutches}, \includegraphics[width=0.35cm]{icon_walker} },
xtick = data, 
enlarge x limits = 0.2,
every x tick label/.append style={font=\tiny} %style={font=\small}
]
\addplot coordinates {(Pedestrian, 7608)(Wheelchair, 6242)(Push_wheelchair, 3323)(Crutches, 5374)(Walker, 7560)};
%\legend{MA}
\end{axis}
\end{tikzpicture}
\caption{Number of instances per class in our dataset.}
\label{fig:numInstances}
\end{figure}

The dataset is subdivided into subsets for training, validation, and evaluation of the pipeline. We use two test sets to evaluate the performance of the pipeline. Test set 1 is used to evaluate the Fast R-CNN detector, whereas Test set 2 is used for the overall evaluation of the pipeline and its ground truth labels consider also occluded objects. Test set 1 is used also as validation set to design the hidden Markov model (HMM) and the tracking algorithm. Fig.~\ref{fig:numInstances} shows the number of instances for each class contained in the set.

\begin{table}[t]
\caption{Object detection performance in terms of mean average precision (MAP) using solely Fast R-CNN.}
\label{tab:FRCNN_performance}
\begin{center}
\begin{tabular}{cccccc|c}
 & \includegraphics[width=0.35cm]{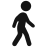} & 
 \includegraphics[width=0.35cm]{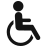} & 
 \includegraphics[width=0.35cm]{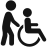} &
 \includegraphics[width=0.35cm]{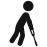} &
 \includegraphics[width=0.35cm]{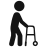} &
 MAP \\
\hline
Ours & 73.0 & 89.6 & 39.5 & 66.9 & 54.9 & \textbf{64.98}\\   
DSW & 70.6 & 86.9 & 42.0 & 69.4 & 49.5 & \textbf{63.67} \\

\hline
\end{tabular}
\end{center}
\end{table}

\begin{table}[t]
\caption{Performance on InOutDoorPeople~\cite{mees2015choosing} test sequence 4. 
%We compare against a dense sliding window approach~\cite{mees2015choosing} and an upper-body detector~\cite{jafari2014real}
}
\label{tab:InOutDoor_performance}
\begin{center}
\begin{tabular}{c|c}
& AP \\
\hline
Ours & 70.0 \\   
DSW~\cite{mees2015choosing} & \textbf{71.6} \\
Upper-body detector~\cite{jafari2014real} & 69.1  \\
\hline
\end{tabular}
\end{center}
\end{table}

\subsection{Detection using Fast R-CNN}

In order to evaluate the object detection performance, we use the standard mean average precision (MAP) metric. We compare the final detection performance using proposals from a dense sliding window method against our fast depth-based region proposal method. In order to create the set of dense sliding window proposals, we applied our templates on the implementation of \citet{mees2015choosing}. 
Tab.~\ref{tab:FRCNN_performance} compares both approaches at an IoU of $0.5$. Our depth-based proposal method performs better than the sliding window baseline method.

\begin{figure*}[]
\centering
\includegraphics[width=\textwidth]{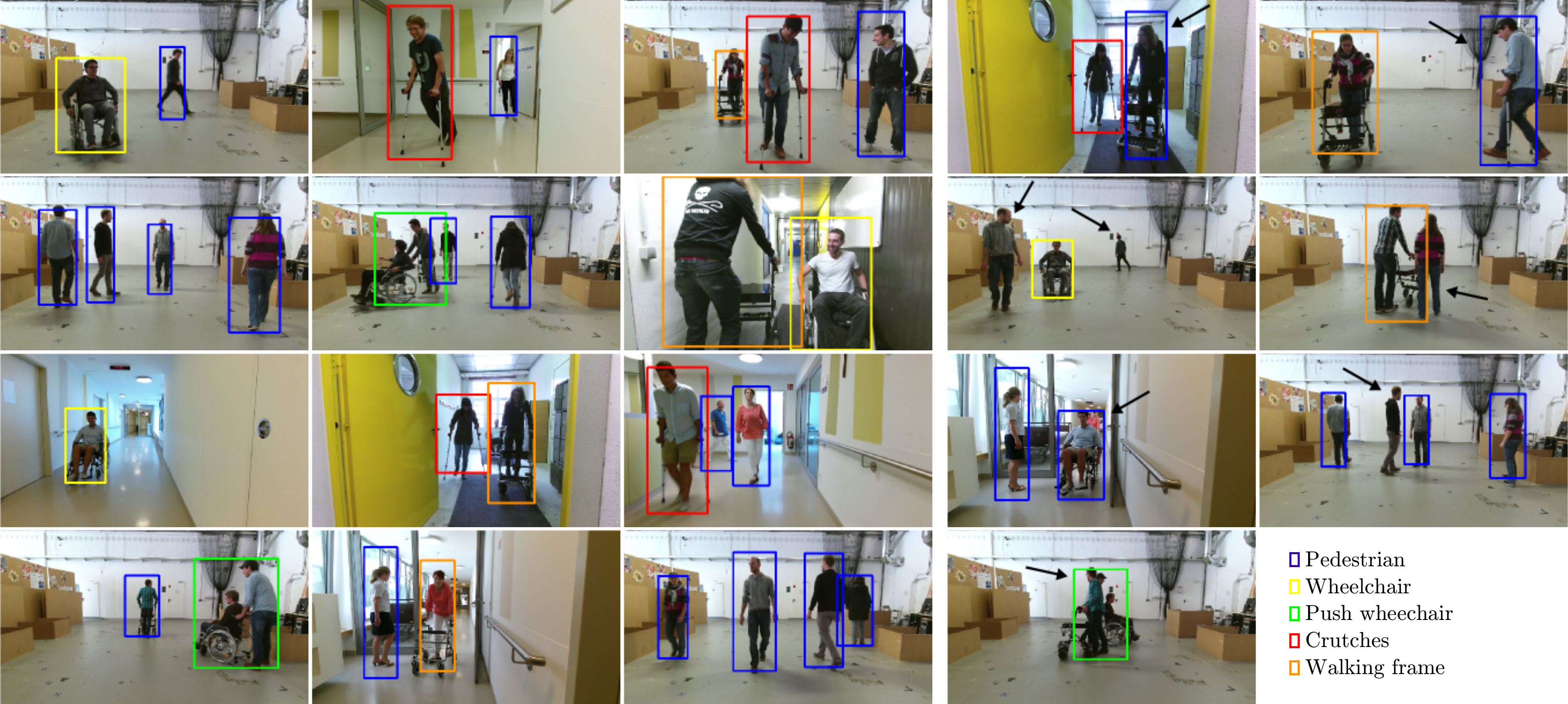}\\
\caption{Qualitative object detection results obtained using our trained Fast R-CNN network. Left: positive examples. Right: cases of failure with missed detections and wrong classifications. We show the RGB image only for demonstration purposes.}
\label{fig:qualitative}
\end{figure*}

We also compare the runtime of both methods on a computer with 12-Core CPU and a GeForce GTX TITAN X with 12GB of memory. Following the dense sliding window algorithm, the set of ROIs for a single image contains approximately 29,000 bounding boxes, while our approach generates an average of 450 proposals. By using dense sliding window (DSW), a single frame is classified in 297 ms, whereas our algorithm requires 43 ms. Our approach is therefore a better choice for employment on a real-world system. We also evaluated the classification performance by means of a confusion matrix where the three most noticeable confusions correspond to $26.8\%$ of people using crutches that were confused with people pushing other people in wheelchairs, $31.7\%$ of people using walking frames that were classified as pedestrians and $23.8\%$ of people using crutches classified as pedestrians. Qualitative detection results are shown in Fig.~\ref{fig:qualitative}.

\begin{figure}
\centering
\begin{tikzpicture}
\begin{axis}[
legend columns = -1,
legend style={
at={(0.5,1.03)},
anchor=south},
ybar, 
ymin=40,
width = 0.8\columnwidth,
height = 0.48\columnwidth,
bar width = 10pt,
ylabel = {Performance ($\%$)},
ylabel near ticks,
%nodes near coords,
symbolic x coords={Fast R-CNN, Fast R-CNN + KF, Fast R-CNN + KF + HMM}, 
xticklabels = {Fast R-CNN, Fast R-CNN \\ + KF, Fast R-CNN \\+ KF + HMM},
xticklabel style={align=center},
xtick = data, 
enlarge x limits = 0.2,
every x tick label/.append style={font=\small} %style={font=\small}
]
%\addplot[fill=blue] coordinates {(Classification, 98.5)(Tracking, 82.32)(Class Estimation, 75.4)};
%\addplot[fill=red] coordinates {(Classification, 80.1)(Tracking, 85.1)(Class Estimation, 82.2)};
%\addplot[fill=green] coordinates {(Classification, 80.6)(Tracking, 72)(Class Estimation, 94)};
%\addplot coordinates {(Classification, 98.5)(Tracking, 82.32)(Class Estimation, 75.4)};
%\addplot coordinates {(Classification, 80.1)(Tracking, 85.1)(Class Estimation, 82.2)};
%\addplot coordinates {(Classification, 80.6)(Tracking, 72)(Class Estimation, 94)};

%results from long nigth
%\addplot coordinates {(Fast R-CNN, 47.37)(Fast R-CNN + KF, 47.80)(Fast R-CNN + KF + HMM, 48.61)};
%\addplot coordinates {(Fast R-CNN, 64.77)(Fast R-CNN + KF, 67.57)(Fast R-CNN + KF + HMM, 70.12)};

%last resuts
\addplot coordinates {(Fast R-CNN, 47.37)(Fast R-CNN + KF, 47.80)(Fast R-CNN + KF + HMM, 51.98)};
\addplot coordinates {(Fast R-CNN, 64.77)(Fast R-CNN + KF, 67.57)(Fast R-CNN + KF + HMM, 73.33)};
\legend{MAP, Recall}
%\legend{Precision, Recall, F1}
\end{axis}
\end{tikzpicture}
\caption{Object detection performance evolution with respect to stand-alone Fast R-CNN. Addition of the two modules KF and HMM to the Fast R-CNN baseline improves overall MAP and recall.}
\label{fig:chart}
\end{figure}
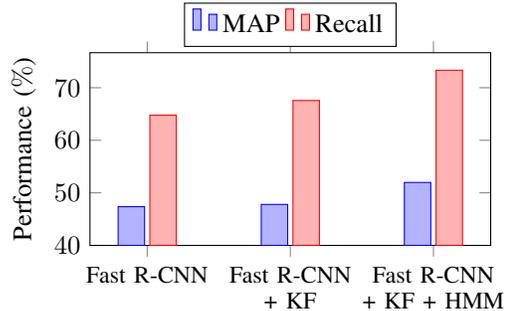
Further, we compare our detector on the InOutDoorPeople dataset~\cite{mees2015choosing} against a dense sliding window approach and an implementation~\cite{dondrup2015} of the depth-based upper-body detector by~\citet{jafari2014real}. 
This dataset only contains labels for pedestrians. 
%We use it to provide a fair comparison with the two baselines, which are not designed to detect multiple classes. 
Therefore, all class predictions of our network are counted as detections of pedestrians.
Table~\ref{tab:InOutDoor_performance} shows that our approach outperforms the upper-body detector but achieves slightly worse results than the DSW baseline. We hypothesize that our detector would improve if trained solely on pedestrians. For all experiments we use the same network that we trained on five classes, without retraining.  

\subsection{Multi-Tracking using Kalman Filter}
To evaluate the performance of the tracking algorithm we use the CLEAR MOT metrics proposed by \citet{bernardin2006multiple} which considers the Multiple Object Tracking Precision (MOTP) and the Multiple Object Tracking Accuracy (MOTA). We obtained a MOTP of $70.60\%$ and $62.19\%$ of MOTA.

\subsection{Framework Evaluation}
We further aim to evaluate our complete pipeline in order to assess the contribution of different stages such as classification, tracking and class estimation in the overall detection performance. In this experiment, we challenge the overall system to estimate the position and class of temporarily occluded people. % by exploiting the predictions of the tracking algorithm. 
We compare the performance of the pipeline in terms of MAP and recall. %, and recall computed as $\text{Recall} = TP/GT$ where $TP$ is the number of true positives and $GT$ the number of ground truth objects.
Results were obtained using our Test set 2 that contains four sequences. Each sequence was evaluated independently and at each frame we use Intersection over Union and the Hungarian algorithm to find ground truth detection pairs. Hypotheses outside of the field of view of the camera were not considered. Test set 2 is especially challenging because of occlusions, which explains the drop in MAP compared to Test set 1.
As can be seen in Fig.~\ref{fig:chart}, the addition of the Kalman filter and the HMM improves the performance of the overall pipeline. The MAP improves by $4.6\%$ from $47.37\%$ to $51.98\%$ compared to the raw Fast R-CNN output. Recall increases from $64.77\%$ to $73.33\%$.   %-------------------------------------------
%--------------   NEW  ---------------------
%-------------------------------------------

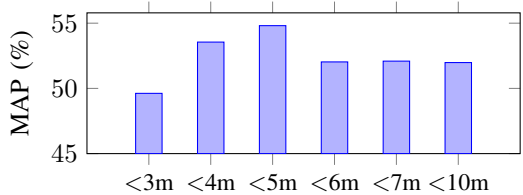
\begin{figure}
\centering
\begin{tikzpicture}
\begin{axis}[
legend columns = -1,
legend style={
at={(0.5,1.03)},
anchor=south},
ybar, 
ymin=45,
width = 0.85\columnwidth,
height = 0.40\columnwidth,
bar width = 10pt,
ylabel = {MAP (\%)},
ylabel near ticks,
%nodes near coords,
symbolic x coords={<three, <four, <five, <six, <seven, All}, 
xticklabels={$<$3m, $<$4m, $<$5m, $<$6m, $<$7m, $<$10m },
xtick = data, 
enlarge x limits = 0.2,
every x tick label/.append style={font=\small} %style={font=\tiny} 
]
% \addplot coordinates {(<three, 49.62)(<four, 53.55)(<five, 54.81)(<six, 52.03)(<seven, 52.09)(<eight, 51.93)(<nine, 52.02)(All, 51.98)};
\addplot coordinates {(<three, 49.62)(<four, 53.55)(<five, 54.81)(<six, 52.03)(<seven, 52.09)(All, 51.98)};
%\legend{MA}
\end{axis}
\end{tikzpicture}
\caption{Variation of the performance with respect to the distance.}
\label{fig:distanceEvaluation}
\end{figure}

We also assess the variation of the performance with respect to the distance from the sensor. For a given distance $d$, this experiment considers detections and ground truth bounding boxes up to $d$ meters only. Fig.~\ref{fig:distanceEvaluation} shows that our method performs best when detecting people up to five meters, achieving an MAP of $54.81\%$. The performance decreases to $51.98\%$ MAP for larger distances and also notably for very short distances.

\begin{figure*}[t]
\centering
\includegraphics[width=0.99\textwidth]{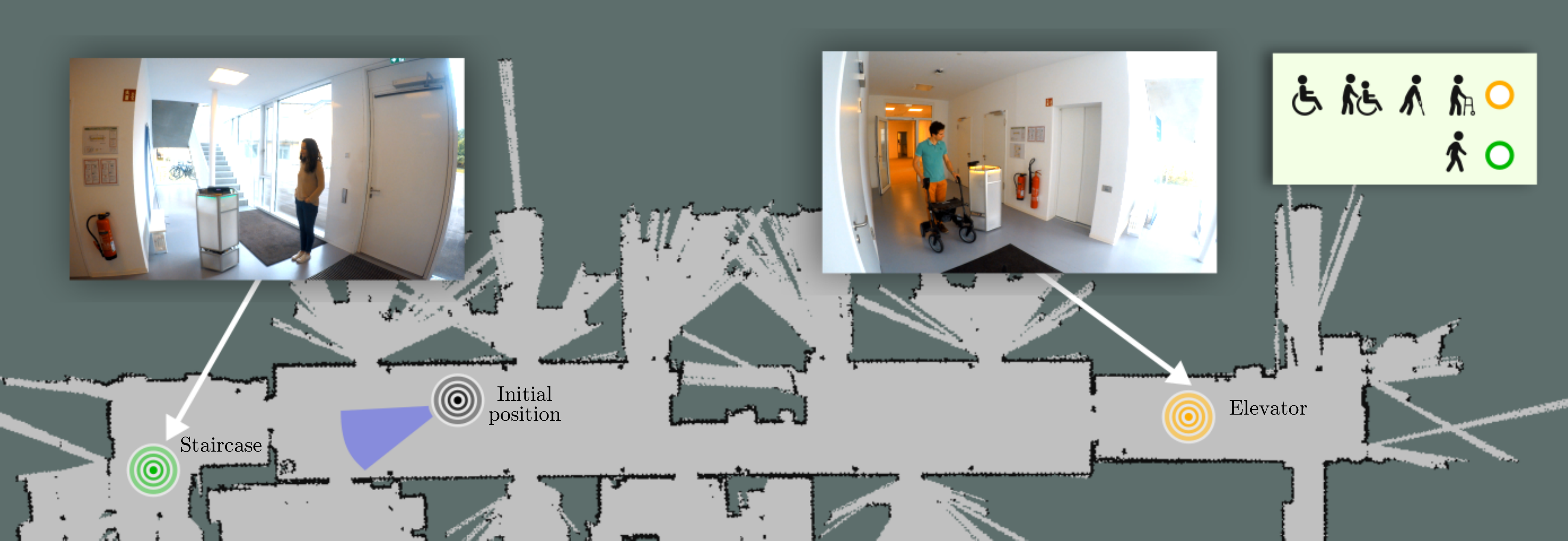}\\
\caption{We use our framework to provide assistance in a person guidance experiment. The robot's task is to guide all pedestrians to the nearest staircase (left image) and all people with mobility aids to the elevator (right image). We told the test subjects to approach the robot at the initial position and then follow the robot's predefined speech commands.}
\label{fig:Realexperiment}
\end{figure*}

%\addtolength{\textheight}{-0.05cm}   % This command serves to balance the column lengths
                                  % the textheight of the last page by a sui amount.

\subsection{Person Guidance Scenario}
To show the applicability of our framework to a real-world service robot task, we test our system in a person guidance scenario. The  robot's task is to guide visitors to the professor's office in our lab, building 80 at the Faculty of Engineering of the University of Freiburg. The professor's office is located at the first floor, opposite the staircase at the main entrance. An elevator is available at the other side of the corridor for people with walking impairments.% who cannot climb stairs. %The robot's task is to greet visitors in the hallway close to the main entrance and guide them either to the stairs or to the elevator, based on the perceived walking capabilities. %Furthermore, the robot is requested to adjust its velocity based on the walking capabilities of visitors.

Our robot uses a laptop computer with an 8-Core CPU and a GeForce GTX 1080 with 8GB of memory in order to process data at approximately 15 frames per second. We use the ROS navigation stack\footnote{\url{http://wiki.ros.org/navigation}} for the navigation parts of the experiment. We select the initial waiting pose of the robot close to the main entrance as well as two goal poses, see Fig.~\ref{fig:Realexperiment}. %The robot uses the positions, velocities and categories of visitors estimated by our pipeline to decide where to guide them. 
The robot observes an area of interest \SI{3}{\meter} in front of it and within \SI{\pm20}{\degree} from its center. Once it detects a person in this area, it waits for \SI{4}{\second} and until the confidence for one category exceeds \SI{90}{\percent} before navigating to one of the two goals. For pedestrians, it navigates to the goal by the stairs; people with mobility aids are guided to the elevator. In addition, the robot adjusts its velocity according to the perceived category of the person. Once it reached its goal, it returns to the waiting position and waits for the next visitor. The robot further uses predefined speech commands to ask the visitors to follow it and inform them how to proceed to the professor's office once its navigation goal is reached.  

We tested thirteen guidance runs with different people from our lab. Each category was tested twice, except pedestrian for which we tested five runs. In all of the runs, the robot perceived the correct category and successfully navigated the person to the correct location. However, in some runs, the people had to adjust their positions to trigger the navigation, because they were too close to the robot or outside of the observed area. The experiment confirms that our approach can be applied on a real robot. Further, it shows how our framework can be used to give appropriate assistance to people, according to their needs.

\section{CONCLUSIONS}

We proposed a perception system to detect and distinguish people according to the mobility aids they use, based on a deep neural network and supported by tracking and class estimation modules. Our experiments show an increase in performance by the addition of these two modules to the pipeline. Moreover, we demonstrated that our approach for the proposal generation can speed up the classification process by a factor of up to seven compared to a dense sliding window baseline while achieving better performance. Additionally, we introduce an RGB-D dataset containing over 17,000 annotated images. In our person guidance experiment we showed that our detection pipeline enables robots to provide individual assistance to people with advanced needs. In the future, we plan to use sensor data from multiple sources to improve our pipeline such as using laser range finder readings to increase the accuracy of the proposal generation at larger distances. We will further examine how the additional information provided by our framework can improve the behavior of robots in populated environments, for example during autonomous navigation.
%or for guiding a person.

\section*{Acknowlegments}
We would like to thank Oier Mees
for his help with the detection experiments.

% section is not required. Although a conclusion may review the main points of the paper, do not replicate the abstract as the conclusion. A conclusion might elaborate on the importance of the work or suggest applications and extensions. 


\bibliographystyle{IEEEtranN}
%\bibliography{IEEEabrv,IEEEexample}
\bibliography{IEEEabrv,AVliterature}

% Generated by IEEEtranN.bst, version: 1.14 (2015/08/26)
\begin{thebibliography}{21}
\providecommand{\natexlab}[1]{#1}
\providecommand{\url}[1]{#1}
\csname url@samestyle\endcsname
\providecommand{\newblock}{\relax}
\providecommand{\bibinfo}[2]{#2}
\providecommand{\BIBentrySTDinterwordspacing}{\spaceskip=0pt\relax}
\providecommand{\BIBentryALTinterwordstretchfactor}{4}
\providecommand{\BIBentryALTinterwordspacing}{\spaceskip=\fontdimen2\font plus
\BIBentryALTinterwordstretchfactor\fontdimen3\font minus
  \fontdimen4\font\relax}
\providecommand{\BIBforeignlanguage}[2]{{%
\expandafter\ifx\csname l@#1\endcsname\relax
\typeout{** WARNING: IEEEtranN.bst: No hyphenation pattern has been}%
\typeout{** loaded for the language `#1'. Using the pattern for}%
\typeout{** the default language instead.}%
\else
\language=\csname l@#1\endcsname
\fi
#2}}
\providecommand{\BIBdecl}{\relax}
\BIBdecl

\bibitem[Girshick(2015)]{girshick2015fast}
R.~Girshick, ``Fast {R-CNN},'' \emph{Int. Conf. on Computer Vision (ICCV)},
  2015.

\bibitem[Ess et~al.(2009)Ess, Leibe, Schindler, and van Gool]{ess2009}
A.~Ess, B.~Leibe, K.~Schindler, and L.~van Gool, ``Robust multiperson tracking
  from a mobile platform,'' \emph{IEEE Transactions on Pattern Analysis and
  Machine Intelligence (TPAMI)}, 2009.

\bibitem[Choi et~al.(2013)Choi, Pantofaru, and Savarese]{choi2013}
W.~Choi, C.~Pantofaru, and S.~Savarese, ``A general framework for tracking
  multiple people from a moving camera,'' \emph{IEEE Transactions on Pattern
  Analysis and Machine Intelligence (TPAMI)}, 2013.

\bibitem[Linder et~al.(2016)Linder, Breuers, Leibe, and Arras]{linder2016}
T.~Linder, S.~Breuers, B.~Leibe, and K.~O. Arras, ``On multi-modal people
  tracking from mobile platforms in very crowded and dynamic environments,'' in
  \emph{IEEE Int. Conf. on Robotics and Automation (ICRA)}, 2016.

\bibitem[Dondrup et~al.(2015)Dondrup, Bellotto, Jovan, and
  Hanheide]{dondrup2015}
C.~Dondrup, N.~Bellotto, F.~Jovan, and M.~Hanheide, ``Real-time multisensor
  people tracking for human-robot spatial interaction,'' in \emph{Int. Conf. on
  Robotics and Automation (ICRA)}, 2015.

\bibitem[Ren et~al.(2015)Ren, He, Girshick, and Sun]{renNIPS15fasterrcnn}
S.~Ren, K.~He, R.~Girshick, and J.~Sun, ``Faster {R-CNN}: Towards real-time
  object detection with region proposal networks,'' in \emph{Advances in Neural
  Information Processing Systems ({NIPS})}, 2015.

\bibitem[Li et~al.(2016)Li, He, Sun, et~al.]{li2016r}
Y.~Li, K.~He, J.~Sun \emph{et~al.}, ``{R-FCN}: Object detection via
  region-based fully convolutional networks,'' in \emph{Advances in Neural
  Information Processing Systems (NIPS)}, 2016.

\bibitem[Redmon and Farhadi(2016)]{redmon2016yolo9000}
J.~Redmon and A.~Farhadi, ``{YOLO}9000: Better, faster, stronger,'' \emph{arXiv
  preprint arXiv:1612.08242}, 2016.

\bibitem[Beyer et~al.(2017)Beyer, Hermans, and Leibe]{beyer2017drow}
L.~Beyer, A.~Hermans, and B.~Leibe, ``{DROW}: Real-time deep learning-based
  wheelchair detection in {2D} range data,'' \emph{IEEE Robotics and Automation
  Letters}, 2017.

\bibitem[Linder et~al.(2015)Linder, Wehner, and Arras]{linder2015real}
T.~Linder, S.~Wehner, and K.~O. Arras, ``Real-time full-body human gender
  recognition in ({RGB})-{D} data,'' in \emph{IEEE Int. Conf. on Robotics and
  Automation (ICRA)}, 2015.

\bibitem[Munaro and Menegatti(2014)]{munaro2014}
M.~Munaro and E.~Menegatti, ``Fast {RGB-D} people tracking for service
  robots,'' \emph{Autonomous Robots}, 2014.

\bibitem[Spinello et~al.(2008)Spinello, Triebel, and
  Siegwart]{spinello2008multimodal}
L.~Spinello, R.~Triebel, and R.~Siegwart, ``Multimodal detection and tracking
  of pedestrians in urban environments with explicit ground plane extraction,''
  in \emph{IEEE/RSJ Int. Conf. on Intelligent Robots and Systems (IROS)}, 2008.

\bibitem[Sudowe and Leibe(2011)]{sudowe2011efficient}
P.~Sudowe and B.~Leibe, ``Efficient use of geometric constraints for
  sliding-window object detection in video,'' \emph{Int. Conf. on Computer
  Vision Systems (ICVS)}, 2011.

\bibitem[Jafari et~al.(2014)Jafari, Mitzel, and Leibe]{jafari2014real}
O.~Jafari, D.~Mitzel, and B.~Leibe, ``Real-time {RGB-D} based people detection
  and tracking for mobile robots and head-worn cameras,'' \emph{Int. Conf. on
  Robotics and Automation (ICRA)}, 2014.

\bibitem[Spinello and Arras(2011)]{spinello2011people}
L.~Spinello and K.~O. Arras, ``People detection in {RGB-D} data,''
  \emph{IEEE/RSJ Int. Conf. on Intelligent Robots and Systems (IROS)}, 2011.

\bibitem[Chen et~al.(2015)Chen, Kundu, Zhu, Berneshawi, Ma, Fidler, and
  Urtasun]{chen20153d}
X.~Chen, K.~Kundu, Y.~Zhu, A.~Berneshawi, H.~Ma, S.~Fidler, and R.~Urtasun,
  ``{3D} object proposals for accurate object class detection,'' \emph{Advances
  in Neural Information Processing Systems (NIPS)}, 2015.

\bibitem[Sharma and Jurie(2011)]{sharma2011learning}
G.~Sharma and F.~Jurie, ``Learning discriminative spatial representation for
  image classification,'' in \emph{British Machine Vision Conf. (BMVC)}, 2011.

\bibitem[Bourdev et~al.(2011)Bourdev, Maji, and Malik]{bourdev2011describing}
L.~Bourdev, S.~Maji, and J.~Malik, ``Describing people: A poselet-based
  approach to attribute classification,'' in \emph{IEEE Int. Conf. on Computer
  Vision (ICCV)}, 2011.

\bibitem[Sudowe et~al.(2015)Sudowe, Spitzer, and Leibe]{sudowe2015person}
P.~Sudowe, H.~Spitzer, and B.~Leibe, ``Person attribute recognition with a
  jointly-trained holistic {CNN} model,'' in \emph{IEEE Int. Conf. on Computer
  Vision Workshops (ICCVW)}, 2015.

\bibitem[Mees et~al.(2016)Mees, Eitel, and Burgard]{mees2015choosing}
O.~Mees, A.~Eitel, and W.~Burgard, ``Choosing smartly: Adaptive multimodal
  fusion for object detection in changing environments,'' \emph{IEEE Int. Conf.
  on Intelligent Robots and Systems (IROS)}, 2016.

\bibitem[Bernardin et~al.(2006)Bernardin, Elbs, and
  Stiefelhagen]{bernardin2006multiple}
K.~Bernardin, E.~Elbs, and R.~Stiefelhagen, ``Multiple object tracking
  performance metrics and evaluation in a smart room environment,'' \emph{IEEE
  Int. Workshop on Visual Surveillance}, 2006.

\end{thebibliography}

\end{document}